\documentclass{article}

\usepackage[square,sort,comma,numbers]{natbib}

\usepackage[preprint]{neurips_2020}

\usepackage[utf8]{inputenc} 
\usepackage[T1]{fontenc}    
\usepackage{hyperref}       
\usepackage{url}            
\usepackage{booktabs}       
\usepackage{amsfonts}       
\usepackage{nicefrac}       
\usepackage{microtype}      
\usepackage[para,online,flushleft]{threeparttable} 
\usepackage{multirow}
\usepackage{amssymb}
\usepackage{mathtools}
\usepackage{pifont}
\usepackage[dvipsnames]{xcolor}
\usepackage{gensymb}

\usepackage{xcolor}

 \usepackage{titlesec}
 
 \titlespacing\section{0pt}{0pt plus 4pt minus 2pt}{0pt plus 2pt minus 2pt}
 \titlespacing\subsection{0pt}{0pt plus 4pt minus 2pt}{0pt plus 2pt minus 2pt}
 \titlespacing\subsubsection{0pt}{0pt plus 4pt minus 2pt}{0pt plus 2pt minus 2pt}

\title{Spatiotemporal Attention for Multivariate Time Series Prediction and Interpretation}

\author{
  Tryambak Gangopadhyay\\
  Department of Mechanical Engineering\\
  Iowa State University\\
  Ames, IA 50011 \\
  \texttt{tryambak@iastate.edu} \\
  \And
  Sin Yong Tan\\
  Department of Mechanical Engineering\\
  Iowa State University\\
  Ames, IA 50011 \\
  \texttt{tsyong98@iastate.edu} \\
  \And
  Zhanhong Jiang\\
  Data Sciences \& Product Development\\
  Johnson Controls\\
  Milwaukee, WI 53202 \\
  \texttt{zhanhong.jiang@jci.com} \\
  \And
  Rui Meng\\
  Lawrence Berkeley National Lab\\
  University of California, Berkeley\\
  Berkeley, CA 94720\\
  \texttt{rmeng@lbl.gov}
  \And
  Soumik Sarkar
  \thanks{This work was supported in part by U.S. AFOSR YIP grant FA9550-17-1-0220 and Iowa State Plant Science Institute Faculty Scholar award.}\\
  Department of Mechanical Engineering\\
  Iowa State University\\
  Ames, IA 50011 \\
  \texttt{soumiks@iastate.edu} \\  
}

\begin{document}

\maketitle

\begin{abstract}
Multivariate time series modeling and prediction problems are abundant in many machine learning application domains. Accurate interpretation of such prediction outcomes from a machine learning model that explicitly captures temporal correlations can significantly benefit the domain experts. In this context, temporal attention has been successfully applied to isolate the important time steps for the input time series. However, in multivariate time series problems, spatial interpretation is also critical to understand the contributions of different variables on the model outputs. We propose a novel deep learning architecture, called spatiotemporal attention mechanism (STAM) for simultaneous learning of the most important time steps and variables. STAM is a \textit{causal} (i.e., only depends on past inputs and does not use future inputs) and \textit{scalable} (i.e., scales well with an increase in the number of variables) approach that is comparable to the state-of-the-art models in terms of computational tractability. We demonstrate our models' performance on two popular public datasets and a domain-specific dataset. When compared with the baseline models, the results show that STAM maintains state-of-the-art prediction accuracy while offering the benefit of accurate spatiotemporal interpretability.
The learned attention weights are validated from a domain knowledge perspective for these real-world datasets.
\end{abstract}

\section{Introduction}
Multivariate time series analysis, classification, and prediction capabilities are crucial for applications in different domains such as healthcare~\cite{bahadori2019temporal}, financial markets~\cite{d2019trimmed}, climate science~\cite{mudelsee2019trend,abatzoglou2020multivariate} and performance monitoring of engineering systems~\cite{gonzalez2019methodology,zhang2019deep}.
Along with the accuracy of decision-making, interpretability remains one of the important aspects of many real-life problems to build user trust and generate domain insights. Unfortunately, however, we often find a trade-off between the model complexity, accuracy, and the ability to interpret the outcomes. Therefore, accurate predictive modeling of multivariate time series coupled with interpretability mechanisms is still a hard technical challenge for the machine learning community.

Long Short Term Memory (LSTM) networks can capture the long-term temporal dependencies in complex multivariate time series \cite{malhotra2015long} and have been used for a variety of applications \cite{shook2018integrating, hua2019deep, gangopadhyay2020deep}.
In Encoder-Decoder model \cite{cho2014learning, sutskever2014sequence}, the information from all the input time-steps is encoded into a single fixed-length vector, which is then used for decoding. 
Deep LSTM based encoder-decoder approach has been developed for multivariate time series forecasting problem without consideration of interpretability \cite{sagheer2019unsupervised}.
To address the bottleneck of using a fixed-length vector in encoder-decoder, a model based on attention was introduced, which can automatically soft search for important parts of the input sequence in neural machine translation \cite{bahdanau2014neural}.
Inspired by this paper, attention mechanism based models have been developed for time series prediction \cite{choi2016retain, qin2017dual, singh2017attend, song2018attend, gangopadhyay2018temporal, zhang2019lstm, li2019ea, liu2020dstp, shook2020crop}.
We compare and contrast some of the notable works in Table \ref{table:findings}. Some of the models are not causal; some are non-scalable and computationally intractable for a large number of variables and with complicated interpretation mechanisms. Also, some models are only meant for single time-step prediction. Among the few spatiotemporal interpretability based approaches, no domain knowledge verification has been provided before. Also, previously developed approaches do not have any spatial attention to align directly with the output that hinders the ability to explicitly capture spatial correlations.

\begin{table*}[ht]
\caption{\textit{Comparisons between existing and proposed mechanisms.}}
\begin{center}
\begin{threeparttable}
\begin{tabular}{c c c c c c}
    \toprule
    \textbf{Method} & \textbf{Causal} & \textbf{Spa.Tem.Int.} & \textbf{Com.Tra.} & \textbf{Seq.Out.}\\ \midrule
      RETAIN~\cite{choi2016retain}&\color{red}\ding{55}&\color{green}\ding{51}&\color{green}\ding{51}&\color{red}\ding{55}\\
      DA-RNN~\cite{qin2017dual}&\color{red}\ding{55}&\color{green}\ding{51}&\color{green}\ding{51}&\color{red}\ding{55}\\
      AttentiveChrome~\cite{singh2017attend}&\color{red}\ding{55}&\color{green}\ding{51}&\color{red}\ding{55}&\color{red}\ding{55}\\
      SAnD~\cite{song2018attend}&\color{red}\ding{55}&\color{red}\ding{55}&\color{green}\ding{51}&\color{red}\ding{55}\\
      ICAtt~\cite{ismail2019input}&\color{green}\ding{51}&\color{red}\ding{55}&\color{green}\ding{51}&\color{red}\ding{55}\\
      DSTP-RNN~\cite{liu2020dstp}&\color{red}\ding{55}&\color{green}\ding{51}&\color{green}\ding{51}&\color{green}\ding{51}\\
      \textbf{STAM} (ours)&\color{green}\ding{51}&\color{green}\ding{51}&\color{green}\ding{51}&\color{green}\ding{51}\\
      \textbf{STAM-Lite} (ours)&\color{green}\ding{51}&\color{green}\ding{51}&\color{green}\ding{51}&\color{green}\ding{51}\\
      \bottomrule
\end{tabular}
\begin{tablenotes}
\item[1] \textbf{Spa.Tem.Int.}: Spatiotemporal Interpretibility; \item[2] \textbf{Com.Tra.}: Computationally Tractable (Scalable); \item[3] \textbf{Seq.Out.}: Sequence Output
\end{tablenotes}
\end{threeparttable}
\end{center}
\label{table:findings}
\end{table*}

To address these limitations, we propose a novel spatiotemporal attention mechanism (STAM) for multivariate time-series prediction that provides meaningful spatiotemporal interpretations. The most important distinguishing features of our approach compared to the previous studies \cite{choi2016retain, qin2017dual, singh2017attend, liu2020dstp} are the causal nature and that both the spatial and temporal attention are aligned directly to the output variable. The model learns the temporal dependencies in data using LSTM layers in the encoder. At each output time-step, the model first soft searches for the most relevant time-steps and most significant variables. The model then predicts based on the computed spatial and temporal context vectors. For multivariate time-series, STAM can be applied in both regression and classification tasks with very minor modifications. The spatial and temporal interpretations can help users better understand the contributions of different features and time-steps for prediction.  

\noindent\textbf{Contributions.}
We summarize the contributions of this work as follows:  


\noindent(1) The proposed STAM architecture is novel for multiple time-step predictions in the context of interpretability for multivariate time-series problems. To the best of our knowledge, this is the first such work on attention-based time series models where the spatial and temporal attention weights are directly aligned to the output, in a causal and scalable (computationally tractable) manner. 

\noindent(2) The spatial and temporal attention mechanisms are jointly trained in a unified architecture to learn the temporal and spatial contributions. The learned interpretations are explained using domain knowledge for real-world datasets. This model can be utilized for any application involving multivariate time series to provide spatiotemporal interpretability for the predictions.

    
\noindent(3) Complexity analysis is provided for STAM. STAM is interpretable while maintaining state-of-the-art prediction accuracy. STAM outperforms the baseline models in most of the experiments, while, for a few experiments, STAM achieves comparable prediction accuracy.

\section{Related Work}
RETAIN \cite{choi2016retain}, based on a two-level neural attention model, detects influential patient visits and significant clinical variables for a binary classification task. In RETAIN, the spatial interpretation method is quite complicated, domain-specific, and only meant for a classification task. 
A dual-stage attention-based recurrent neural network (DA-RNN) \cite{qin2017dual} has spatial attention in the encoder layer and temporal attention in the decoder. DA-RNN is not causal as it also depends on future inputs during the computation of the spatial weights in the encoding phase.
Their work also suffers from a lack of understanding of the attention weights from a domain knowledge perspective. 
AttentiveChrome \cite{singh2017attend} was developed to model and interpret dependencies among chromatin factors by using two levels of attention for a single output prediction (binary classification). The hierarchy of LSTM networks makes AttentiveChrome non-scalable and computationally intractable in the presence of a large number of variables. 
Using bidirectional LSTMs to encode, the model is non-causal for time series domain applications. 
Transformer \cite{vaswani2017attention}, based solely on attention mechanisms, has achieved state-of-the-art results for the machine translation task. However, the Transformer can only highlight sequential attention weights and will not be suitable for spatial interpretability in a causal way for time-series predictions.
For clinical time-series modeling, attention-based model SAnD \cite{song2018attend} has been utilized inspired by the Transformer \cite{vaswani2017attention} model with some architectural modifications. Though computationally tractable, SAnD is not causal and does not provide spatiotemporal interpretability. 
Some attention-based models \cite{zhang2019lstm, li2019ea, ismail2019input} lacks in spatiotemporal interpretability having only spatial attention. 
Non-causal DSTP-RNN model \cite{liu2020dstp} utilize multiple attention layers at each encoding time-step which complicates the interpretation. 
Similar to DA-RNN, DSTP-RNN has spatial attention in the encoding phase, the limitations of which are described in the next section. 
Since the official codebase remains unpublished at the time of this submission, we do not perform a comparative study for DSTP-RNN. 
Other non-causal models developed in this domain include a multi-stage attention network \cite{hu2020multistage} and a bidirectional LSTM network with temporal attention \cite{du2020multivariate}.

\section{Preliminaries}
\label{prelim}

\subsection{Notations and Problem Formulation}
We introduce the notations to be used in this paper and formulate the problem we aim to study.
Given $N$ time series, we denote by $\mathbf{X}=[\mathbf{x}^1,\mathbf{x}^2,...,\mathbf{x}^N]^{\top}\in\mathbb{R}^{N\times T_x}$, the compact form of all time series, where $T_x$ is the total input sequence length and $\mathbf{x}^i=[x^i_1,x^i_2,...,x^i_{T_x}]^{\top}\in\mathbb{R}^{T_x}, i\in\{1,2,...,N\}$ signifies time series associated with each input variable. 
To represent all input variables at time step $t\in\{1,2,...,T_x\}$, with a slight abuse of notation, we denote by $\mathbf{x}_t=[x^1_t,x^2_t,...,x^N_t]^{\top}\in\mathbb{R}^N$ such that the compact form of all time series can also be expressed as $\mathbf{X}=[\mathbf{x}_1,\mathbf{x}_2,...,\mathbf{x}_{T_x}]^{\top}$. 
Analogously, we denote by $\mathbf{y}\in\mathbb{R}^{T_y}$ the output time series for $T_y$ time-steps, where $\mathbf{y}_j\in\mathbb{R}$ is the output at time step $j$.
For future time series prediction problems, given the historical information for $T_x$ input (multivariate) time-steps, an employed sequence model aims at learning a (non)linear mapping for $T_y$ future values of the output (univariate) time series. 
To mathematically formulate this problem, we define $\mathcal{F}(\cdot)$ as the mapping to be learned to obtain the prediction of $\hat{\mathbf{y}}_j$ at output time-step $j$.
\begin{equation}\label{pro_for}
\hat{\mathbf{y}}_j = \mathcal{F}(\hat{\mathbf{y}}_1,\hat{\mathbf{y}}_2,...,\hat{\mathbf{y}}_{j-1},\mathbf{x}_1,\mathbf{x}_2,...,\mathbf{x}_{T_x})
\end{equation}
In this paper, we aim to develop novel mapping functions $\mathcal{F}$ in Eq.~\ref{pro_for} that achieves highly comparable or better prediction accuracy while shedding light on both spatial and temporal relationships between input and output. Compared to the existing works mentioned in the last section, our work facilitates the accurate spatiotemporal interpretability that is quite crucial in time series prediction problems. 

\subsection{Attention Mechanism}
Various attention mechanisms have been proposed and popularly applied to different deep sequence models, such as RNN, GRU, and LSTM~\cite{zhang2018recurrent,wang2016attention}. 
We introduce here the existing attention mechanisms. 
We denote by $\mathbf{h}_{t-1}\in\mathbb{R}^m$ and $\mathbf{c}_{t-1}\in\mathbb{R}^m$ the encoder hidden state and cell state at time $t-1$ respectively. It is well known that $\mathbf{h}_t$ and $\mathbf{c}_t$ can be calculated by leveraging the update laws of LSTM \cite{hochreiter1997long}.

\noindent\textbf{Spatial Attention Mechanism.} 
The \textit{spatial} attention mechanism can determine the relative contributions of different input variables in multivariate time series prediction.
Recently, a number of papers \cite{qin2017dual, liu2020dstp, hu2020multistage} have proposed to incorporate spatial attention in the encoding phase.
Given the $i$-th attribute time series $\mathbf{x}^i$ of length $T_x$, the spatial attention $\beta^i_t$ at time-step $t$ is computed as following.
\begin{equation}\label{spa_att1}
    \begin{split}
        &e^i_t = \mathbf{v}_e^{\top}\textnormal{tanh}(W_e[\mathbf{h}_{t-1};\mathbf{c}_{t-1}]+U_e\mathbf{x}^i)\\
        &\beta^i_t = \frac{\textnormal{exp}(e^i_t)}{\sum_{o=1}^N\textnormal{exp}(e^o_t)}
    \end{split}
\end{equation}
The raw input time series at time $t$, $\mathbf{x}_t$, is then replaced by the $\textit{weighted}$ time series $\hat{\mathbf{x}}_t$, and with $\hat{\mathbf{x}}_t$ as input to the encoder LSTM (function $f_1$), the new states $\mathbf{h}_t$ and $\mathbf{c}_t$ are computed.
\begin{equation}\label{weighted_input1}
    \begin{split}
        \hat{\mathbf{x}}_t=[\beta^1_tx^1_t,\beta^2_tx^2_t,...,\beta^N_tx^N_t]^{\top}\\
        (\mathbf{h}_t,\mathbf{c}_{t})=f_1(\mathbf{h}_{t-1},\mathbf{c}_{t-1},\hat{\mathbf{x}}_t)
    \end{split}
\end{equation}



\noindent\textbf{Temporal Attention Mechanism.} 
The original \textit{temporal} attention mechanism \cite{bahdanau2014neural} was proposed to be used in the decoding phase after the encoder.
At output time-step $j$ of the decoder, the attention weight of each encoder hidden state is calculated by Eq.~\ref{tem_att1}.
\begin{equation}\label{tem_att1}
        \alpha_j^t=
        \frac{\textnormal{exp}(a_j^t)}{\sum_{l=1}^{T_x}\textnormal{exp}(a_j^l)},\;
        \mathbf{s}_j = \sum_{t=1}^{T_x}\alpha_j^t\mathbf{h}_t.
\end{equation}
The probability $\alpha_j^t$ reflects how much the output $\mathbf{y}_j$ is aligned to the input $\mathbf{x}_t$. 
The associated energy $a_j^t$ is computed using an alignment model (feed forward neural network), which is a function of $\mathbf{h}_t\in\mathbb{R}^{m}$ and previous decoder hidden state $\mathbf{h'}_{j-1}\in\mathbb{R}^{p}$. 
The \textit{temporal context} vector $\mathbf{s}_j$ is the input to the decoder at output time-step $j$. 
Intuitively, most temporal interpretability works \cite{qin2017dual, singh2017attend, liu2020dstp, hu2020multistage} have adopted this approach to compute the temporal attention weights.

\noindent\textbf{Limitations.} 
Recent works in multivariate time series prediction \cite{qin2017dual, liu2020dstp, hu2020multistage, yoshimiforecasting} have developed different \textit{spatiotemporal} attention mechanisms by incorporating the spatial attention into the encoder layer followed by temporal attention into the decoder layer, as done previously. Unfortunately, there exist two major limitations in these spatiotemporal attention mechanisms:

\noindent(1) The \textit{causality} is broken by using $\mathbf{x}^i, 1\leq i\leq N$ covering the whole length of $T_x$ to compute the spatial attention weights (Eq.~\ref{spa_att1}) which are used to calculate the $\textit{weighted}$ time series $\hat{\mathbf{x}}_t$ (Eq.~\ref{weighted_input1}) at time-step $t$. The time-step $t$ is ranging from 1 to $T_x$ and for each $t$, the spatial attention weight calculations require future information ahead of $t$. Using $\hat{\mathbf{x}}_t$ as input, the hidden state of the encoder LSTM $\mathbf{h}_t$ is computed, which has implicit future information ahead of time-step $t$. It therefore affects temporal interpretability as well because the temporal attention alignment model is dependent on $\mathbf{h}_t$. 

\noindent(2) There is no such \textit{spatial context} vector to align with the output sequence directly, as in the temporal attention. Although the current approaches measure the spatial importance in multivariate input time series, based on Eq.~\ref{weighted_input1}, the spatial relationships between input and output can only be captured implicitly via the hidden states. Therefore, the existing approaches still lack accurate spatiotemporal interpretability.

\section{Spatiotemporal Attention Mechanism (STAM)}   

We address the two limitations stated above by introducing a novel spatiotemporal attention mechanism (STAM) to (1) maintain the causality in the model and to (2) achieve accurate spatiotemporal interpretability.
In this section, we propose and investigate the model STAM and its computational complexity. 
The STAM model is illustrated in Fig.~\ref{stam_2}.
We also propose a lighter version of STAM, called STAM-Lite in the supplementary materials.

\begin{figure*}
  \centering
    \includegraphics[width=12cm,keepaspectratio]{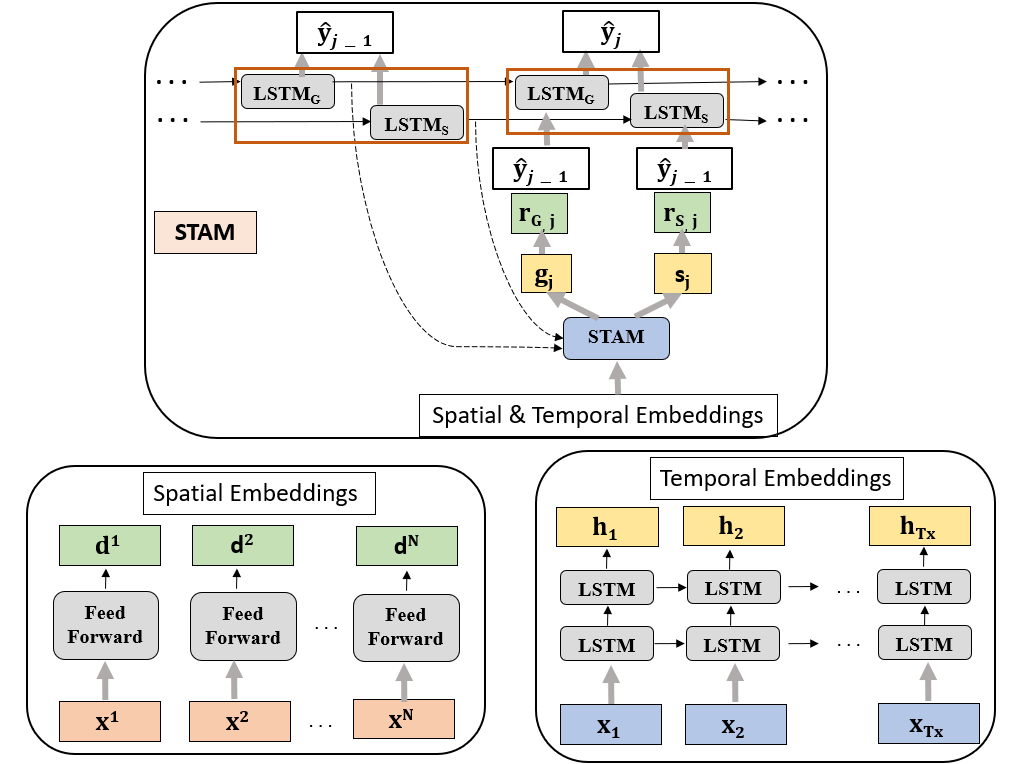} 
  \caption{\textit{The illustration of the proposed model STAM attempting to compute the output $\hat{\mathbf{y}}_j$ at time step $j$.}}
   \vspace{-7pt}
  \label{stam_2}
\end{figure*}

\subsection{Spatial and Temporal Attention}
In STAM, \textit{we develop a spatiotemporal attention mechanism to come up with spatial and temporal context vectors to align directly with the output variable}.
The intuition behind such an idea is that instead of having the spatial attention in the encoder layer, a separate spatial attention is designed parallel to the temporal attention in the decoder layer to simultaneously attend to the most relevant time steps and the most significant variables.
Therefore, in STAM, both the spatial and temporal attentions align directly with the output. 
The inputs to the spatial and temporal attention are spatial and temporal embeddings, respectively.
The embeddings are generated independently.
The spatial embeddings are obtained by using feed forward neural network for each feature $\mathbf{x}^i=[x^i_1,x^i_2,...,x^i_{T_x}]^{\top}\in\mathbb{R}^{T_x}, i\in\{1,2,...,N\}$.
From $\mathbf{X}=[\mathbf{x}^1,\mathbf{x}^2,...,\mathbf{x}^N]^{\top}$, the embeddings for all variables are computed as $\mathbf{D}=[\mathbf{d}^1,\mathbf{d}^2,...,\mathbf{d}^N]^{\top}$, where $\mathbf{d}^{i}\in\mathbb{R}^m$.
Independently, the encoder consisting of two stacked LSTM layers, compute the temporal embeddings (hidden states) given the input time series $\mathbf{X}=[\mathbf{x}_1,\mathbf{x}_2,...,\mathbf{x}_{T_x}]^{\top}$. At time-step $t$, the input to the encoder is  $\mathbf{x}_t=[x^1_t,x^2_t,...,x^N_t]^{\top}\in\mathbb{R}^N$.
After reading the input sequence in order from $\mathbf{x}_1$ to $\mathbf{x}_{T_x}$, the first LSTM layer returns the hidden states which act as inputs to the second LSTM layer of the encoder. 
Using two LSTM layers, the encoder generates a sequence of hidden states, expressed as $\mathbf{H}=[\mathbf{h}_1,\mathbf{h}_2,...,\mathbf{h}_{T_x}]^{\top}$, where $\mathbf{h}_{t}\in\mathbb{R}^m$.

The hidden state and cell state in the decoder layer are denoted by $\mathbf{h'}\in\mathbb{R}^p$ and $\mathbf{c'}\in\mathbb{R}^p$. A feed-forward neural network is used as an alignment model to compute the spatial attention weights. 
At output time-step $j$, the $i$-th spatial attention weight $\beta^i_j$ is calculated, where $[\mathbf{h'}_{j-1};\mathbf{d}^i]\in\mathbb{R}^{p+m}$ with $\mathbf{h'}_{j-1}\in\mathbb{R}^p$ the previous hidden state of the decoder LSTM and $\mathbf{d}^{i}\in\mathbb{R}^m$ the spatial embedding for $i$-th  feature.
The parameters to learn are $W_e\in\mathbb{R}^{p+m}$ and $b_e\in\mathbb{R}$.
We adopt the ReLU activation function instead of tanh due to slightly better results obtained through empirical studies.
We then calculate the spatial context vector $\mathbf{g}_j$ using the spatial attention weights.
\begin{equation}\label{spa_att_stam}
    \begin{split}
        &e^i_j = \textnormal{ReLU}(W_e^{\top}[\mathbf{h'}_{j-1};\mathbf{d}^i] + b_e)\\
        &\beta^i_j = \frac{\textnormal{exp}(e^i_j)}{\sum_{o=1}^N\textnormal{exp}(e^o_j)}, \;
    \mathbf{g}_j = \sum_{i=1}^N\beta^i_{j}\mathbf{d}^i    
    \end{split}
\end{equation}
For output time-step $j$, to get the temporal attention weight $\alpha_j^t$ corresponding to the hidden state $\mathbf{h}_t$, the associated energy $a_j^t$ is computed as follows:
\begin{equation}\label{tem_att_stam}
        a^t_j = \textnormal{ReLU}(W_a^{\top}[\mathbf{h'}_{j-1};\mathbf{h}_t] + b_a)\\
\end{equation}
where $[\mathbf{h'}_{j-1};\mathbf{h}_t]\in\mathbb{R}^{p+m}$ with $\mathbf{h'}_{j-1}\in\mathbb{R}^p$ the previous decoder hidden state and $\mathbf{h}_t\in\mathbb{R}^m$ the temporal embedding for $t$-th input time-step.
The parameters to learn are $W_a\in\mathbb{R}^{p+m}$ and $b_a\in\mathbb{R}$.
Thereafter, the attention weights $\alpha_j^t$ for $t\in\{1,2,...,T_x\}$ are calculated followed by the temporal context vector $\mathbf{s}_j$ according to Eq.~\ref{tem_att1}.
It should be noted that the spatial context vector $\mathbf{g}_j$ and the temporal context vector $\mathbf{s}_j$ are distinct at each time step. 

\subsection{STAM}
In STAM, the spatial and temporal attentions rely on two different LSTM layers (denoted by $\textnormal{LSTM}_G$ and $\textnormal{LSTM}_S$) in the decoder. Therefore, the hidden state inputs are different to spatial and temporal attention.
For the purpose of presentation, we denote the hidden and cell states of the LSTM corresponding to the spatial context $\mathbf{g}_j$ with $*_{G}$, where $*$ indicates either hidden or cell state. The states are therefore denoted as $\mathbf{h'}_{G}\in\mathbb{R}^p$ and $\mathbf{c'}_{G}\in\mathbb{R}^p$.
At output time-step $j$, the input to the spatial attention (Eq.~\ref{spa_att_stam}) is the previous decoder hidden state $\mathbf{h'}_{G,j-1}$ to compute the spatial context vector $\mathbf{g}_j$.
The dimension of $\mathbf{g}_j$ is reduced to $\mathbf{r}_{G,j}$ using a feed-forward neural network. We optimize the extent of this reduction through our experiments. 
Next we update $\mathbf{r}_{G,j}$ by concatenating with output of the previous time-step $\hat{\mathbf{y}}_{j-1}$. The concatenation is denoted by $\hat{\mathbf{r}}_{G,j}$. It should be noted that the decoder output $\hat{\mathbf{y}}_{j-1}$ is a scalar in time series prediction instead of a vector.
$\hat{\mathbf{r}}_{G,j}$ becomes the input to $\textnormal{LSTM}_G$ as follows:
\begin{equation}\label{stam2_concat_spa_1}
    \begin{split}
    &\mathbf{r}_{G,j} = \textnormal{ReLU}(W_{G}\mathbf{g}_j + b_{G}),\;
    \hat{\mathbf{r}}_{G,j} = [\mathbf{r}_{G,j}; \hat{\mathbf{y}}_{j-1}]
    \end{split}
\end{equation}
where $W_{G}\in\mathbb{R}^{q\times m}$ and $b_{G}\in\mathbb{R}^q$ are learnable parameters.
Instead of the real measurement, the prediction of the output time series is utilized through the \textit{non-teacher forcing} training approach, which allows a more robust learning process.
Similarly, for the LSTM corresponding to the temporal context $\mathbf{s}_j$, the states are indicated with $*_{S}$.
The hidden and cell states are denoted by $\mathbf{h'}_{S}\in\mathbb{R}^p$ and $\mathbf{c'}_{S}\in\mathbb{R}^p$ respectively.
At output time-step $j$, the previous $\textnormal{LSTM}_S$ hidden state $\mathbf{h'}_{S,j-1}$ is the input to the temporal attention (Eq.~\ref{tem_att_stam}) to compute the temporal context vector $\mathbf{s}_j$ (Eq.~\ref{tem_att1}).
After reducing the dimension reduction and concatenating with $\hat{\mathbf{y}}_{j-1}$, $\hat{\mathbf{r}}_{S,j}$ becomes the input to $\textnormal{LSTM}_S$ as shown here:
\begin{equation}\label{stam2_concat_spa_2}
    \begin{split}
    &\mathbf{r}_{S,j} = \textnormal{ReLU}(W_{S}\mathbf{s}_j + b_{S}),\;
    \hat{\mathbf{r}}_{S,j} = [\mathbf{r}_{S,j}; \hat{\mathbf{y}}_{j-1}]
    \end{split}
\end{equation}
where $W_{S}\in\mathbb{R}^{q\times m}$ and $b_{S}\in\mathbb{R}^q$ are parameters to learn. The decoder hidden states and cell states are updated as:
\begin{equation}\label{stam_2_dec_hidden}
    \begin{split}
    &(\mathbf{h'}_{G,j}, \mathbf{c'}_{G,j}) = f_3(\mathbf{h'}_{G,j-1}, \mathbf{c'}_{G,j-1}, \hat{\mathbf{r}}_{G,j})\\
    &(\mathbf{h'}_{S,j}, \mathbf{c'}_{S,j}) = f_4(\mathbf{h'}_{S,j-1}, \mathbf{c'}_{S,j-1}, \hat{\mathbf{r}}_{S,j})
    \end{split}
\end{equation}
where $f_3$ and $f_4$ are the nonlinear mappings of $\textnormal{LSTM}_G$ and $\textnormal{LSTM}_S$ respectively. Before prediction, the last step for STAM is to unify the hidden state updates of the two LSTMs together by concatenating into $[\mathbf{h'}_{G,j};\mathbf{h'}_{S,j}]$.
Compared to STAM-Lite, STAM has one extra LSTM layer added to the decoder to separately account for the spatial and temporal impact of input time series on the output or target time series. 
We can observe that the spatial and temporal context vectors are aligned directly with the output variable enabling accurate interpretability, which will be shown empirically.

\subsection{Complexity Analysis}
In a principled manner, STAM mainly involves four modules: the encoder, the spatial attention, the temporal attention, and the decoder. 
Therefore, we omit some operations such as spatial embeddings and concatenation reduction for the context vectors. 
The encoder and decoder in STAM are spanned by LSTM layers. 
STAM has two LSTM layers in the encoder, and its decoder module also uses two LSTM layers separately to process the context vectors generated from the spatial and temporal attention, respectively.
To analyze the STAM's inference time complexity, we follow a similar calculation method as in \cite{tran2018temporal}.     
As the state size of the encoder is $m$, and that of the decoder is $p$, by taking into account all the four modules, such a mechanism leads to the inference time complexity of $\mathcal{O}(8(Nm+m^2+2m)T_x+(p+2+2m)(N+T_x)T_y+8T_y(p^2+pq+3p))$, where $q$ is the dimension of the context vector after dimension reduction for STAM. The first and third terms inside signify the computational complexity for the encoder and decoder, respectively. The second term is for spatial and temporal attention.

\section{Experiments}
\label{exp}

\subsection{Datasets}

\textbf{Pollution Dataset:} 
We use the Beijing PM2.5 Data Set from the UCI Machine Learning Repository. It is an hourly dataset comprising the PM2.5 data of the US Embassy in Beijing and meteorological data from Beijing Capital International Airport \cite{liang2015assessing}.
For $T_x$ input time-steps, we use eight variables - pollution (PM2.5 concentration), dew point, temperature, pressure, combined wind direction, cumulated wind speed, cumulated hours of snow, and cumulated hours of rain. 
We predict the pollution for the upcoming $T_y$ time-steps. 
Keeping the last 20\% of the dataset for testing, the approximate sizes of the training, validation, and test sets are 26,275, 8,758, and 8,759, respectively.

\noindent\textbf{Building Dataset:}
We use a public multivariate time series dataset collected from an air handling unit in a building heating, ventilation, and air-conditioning (HVAC) system \cite{OpenEI}. 
This dataset consists of nine variables - average zone temperature (AvgZoneTemp), outside air temperature (OAT, $\degree F$), return air temperature (RAT, $\degree F$), outside air damper command (OA DamperCMD), cooling valve command (CoolValveCMD), discharge air temperature (DAT, $\degree F$), supply fan speed command (SuFanSpeedCMD), discharge air static pressure (DA StaticP), and return fan speed command (ReFanSpeedCMD). 
With this multivariate input of $T_x$ time-steps, the output is average zone temperature for upcoming $T_y$ time-steps. The training, validation, and test sizes are approximately 20,932, 6,977, and 6,978, respectively.

\noindent\textbf{EHR Dataset:}
MIMIC-III \cite{johnson2016mimic} is a publicly available electronic health record (EHR) database that comprises information relating to patients admitted to critical care units at a large tertiary care hospital.
Clinical prediction benchmark tasks have been proposed using MIMIC-III \cite{harutyunyan2019multitask} containing multivariate time series data.
Instead of using the benchmark tasks, we formulate a new task that can be most suitable to test our proposed model STAM.
After getting access to MIMIC-III, we only followed \cite{harutyunyan2019multitask} to generate the training, validation, and test datasets with each having 14681, 3222, and 3236 number of ICU stays, respectively.
We choose seven clinical variables - Glucose, Heart Rate, Mean Blood Pressure (MBP), Oxygen Saturation ($O_2$sat), Respiratory Rate (RR), Temperature, and pH level. Keeping these variables as input for $T_x = 24$ time-steps, we predict respiratory rate for $T_y$ future time-steps. In other words, based on the clinical variables on the first day (24 hours) of an ICU stay, we predict the respiratory rate for the upcoming $T_y$ hours.

\subsection{Baseline Models and Results}
\label{Baseline Methods and Hyperparameters}

\textbf{Baseline Models:} 
For the empirical results comparison, we use these baseline models: Epsilon-Support Vector Regression with Radial Basis Function kernel (SVR-RBF), Encoder-Decoder (Enc-Dec) model \cite{cho2014learning, sutskever2014sequence}, LSTM with temporal attention (LSTM-Att) model \cite{bahdanau2014neural}, and Dual-Stage Attention-Based Recurrent Neural Network (DA-RNN) \cite{qin2017dual}.
We try to optimize the hyper-parameters of all the baseline models, including SVR-RBF, which have shown improved results than in \cite{du2020multivariate}. The optimized values of hidden state dimensions for the Enc-Dec, LSTM-Att, and DA-RNN models are 32, 32, and 64. With this setting, the approximate number of trainable parameters for Enc-Dec, LSTM-Att and DA-RNN are 18,831, 19,120 and 57,764 respectively.
Additional details of the baseline models are provided in the supplementary materials.

\begin{table*}[ht]
\caption{\textit{Empirical results for \textbf{pollution} dataset (with $T_x=5$, $T_y=4$). Each model was trained three times, to obtain the average and standard deviation of each evaluation metric. 
}}
\begin{center}
\begin{threeparttable}
\begin{tabular}{|l|c|c|c||c|c|}
    \toprule
Model     & RMSE                      & MAE                       & $R^2$ Score                 & Train Time & Test \\ 
 & & & & / epoch & Time \\ \midrule\midrule
SVR-RBF & 48.135 $\pm$ 0.000 & 31.890 $\pm$ 0.000 & 0.735 $\pm$ 0.000 & 11.96s & 2.82s \\ \midrule
Enc-Dec & 48.043 $\pm$ 0.209 & 35.817 $\pm$ 10.587 & 0.736 $\pm$ 0.002 & 6.18s & 0.61s \\ \midrule
LSTM-Att & 47.957 $\pm$ 0.377 & 30.730 $\pm$ 0.448 & 0.737 $\pm$ 0.004 & 6.79s & 0.64s \\ \midrule
DA-RNN  & 49.207 $\pm$ 0.106 & 31.267 $\pm$ 0.220 & 0.723 $\pm$ 0.001 & 7.36s & 0.63s \\ \midrule
STAM-Lite & \textbf{47.658 $\pm$ 0.155} & 30.080 $\pm$ 0.673 & \textbf{0.741 $\pm$ 0.002} & 7.15s & 0.64s \\ \midrule
STAM & 47.778 $\pm$ 0.404 & \textbf{29.853 $\pm$ 0.965} & 0.739 $\pm$ 0.004 & 8.84s & 0.86s \\
\bottomrule
\end{tabular}
\end{threeparttable}
\end{center}
\label{table:Pollution results}
\end{table*}

\begin{table*}[ht]
\caption{\textit{Empirical results for \textbf{building} dataset (with $T_x=5$, $T_y=3$). Each model was trained three times, to obtain the average and standard deviation of each evaluation metric.
}}
\begin{center}
\begin{threeparttable}
\begin{tabular}{|l|c|c|c||c|c|}
    \toprule
Model     & RMSE                      & MAE                       & $R^2$ Score                 & Train Time & Test \\ 
 & & & & / epoch & Time \\ \midrule\midrule
SVR-RBF & 0.1344 $\pm$ 0.0000 & 0.1125 $\pm$ 0.0000 & 0.9874 $\pm$ 0.0000 & 0.19s & 0.03s \\ \midrule
Enc-Dec & 0.0776 $\pm$ 0.0189 & 0.0586 $\pm$ 0.0207 & 0.9956 $\pm$ 0.0022 & 4.40s & 0.42s \\ \midrule
LSTM-Att  & 0.0601 $\pm$ 0.0045 & 0.0450 $\pm$ 0.0039 & \textbf{0.9975 $\pm$ 0.0004} & 4.76s & 0.42s \\ \midrule
DA-RNN  & 0.0600 $\pm$ 0.0013 & 0.0408 $\pm$ 0.0006 & \textbf{0.9975 $\pm$ 0.0002} & 5.18s & 0.48s \\ \midrule
STAM-Lite    & 0.0634 $\pm$ 0.0017 & 0.0448 $\pm$ 0.0002 & 0.9972 $\pm$ 0.0002 & 5.20s & 0.53s \\ \midrule
STAM    & \textbf{0.0599 $\pm$ 0.0024} & \textbf{0.0415 $\pm$ 0.0010} & \textbf{0.9975 $\pm$ 0.0002} & 5.94s & 0.60s \\
      \bottomrule
\end{tabular}
\end{threeparttable}
\end{center}
\label{table:Building results}
\end{table*}

\begin{table*}[ht]
\caption{\textit{Empirical results for \textbf{EHR} dataset (with $T_x=24$, $T_y=4$). Each model was trained three times, to obtain the average and standard deviation of each evaluation metric.
}}
\begin{center}
\begin{threeparttable}
\begin{tabular}{|l|c|c|c||c|c|}
    \toprule
Model     & RMSE                      & MAE                       & $R^2$ Score & Train Time & Test \\ 
 & & & & / epoch & Time \\ \midrule\midrule
SVR-RBF & 4.6067 $\pm$ 0.0000 & 3.2984 $\pm$ 0.0000 & 0.4078 $\pm$ 0.0000 & 26.92s & 5.22s \\ \midrule
Enc-Dec & 4.5981 $\pm$ 0.0088 & 3.2807 $\pm$ 0.0202 & 0.4100 $\pm$ 0.0023 & 6.48s & 0.47s \\ \midrule
LSTM-Att  & 4.6269 $\pm$ 0.0365 & 3.3019 $\pm$ 0.0321 & 0.4026 $\pm$ 0.0094 & 6.77s & 0.53s \\ \midrule
DA-RNN  & 4.6760 $\pm$ 0.0216 & 3.3011 $\pm$ 0.0151 & 0.3898 $\pm$ 0.0056 & 10.64s & 0.63s \\ \midrule
STAM-Lite    & 4.6089 $\pm$ 0.0107 & 3.3085 $\pm$ 0.0141 & 0.4072 $\pm$ 0.0027 & 7.06s & 0.53s \\ \midrule
STAM & \textbf{4.5913 $\pm$ 0.0031} & \textbf{3.2805 $\pm$ 0.0067} & \textbf{0.4117 $\pm$ 0.0008} & 7.93s & 0.56s \\
      \bottomrule
\end{tabular}
\end{threeparttable}
\end{center}
\label{table:EHR results}
\end{table*}

\smallskip
\noindent\textbf{Results:}
We perform experiments to come up with the best set of hyper-parameters for training our STAM model.
Keeping the hidden state dimensions of the encoder and decoder same ($m=p$) for simplicity, a dimension of 32 gives better results in the experiments. We use Adam optimizer with a learning rate of 0.001 and a batch size of 256. 
To prevent overfitting, dropout layer (0.2) is used after each LSTM layer, and each model is trained for 50 epochs.
Through experiments, we optimize the dimension reduction of context vectors to $q=4$ and the input sequence length to $T_x$ = 5 for pollution and building datasets.
Under these settings, the approximate number of trainable parameters for STAM is 24,733.
We use three evaluation metrics: root mean square error (RMSE), mean absolute error (MAE), and coefficient of determination or R-squared score ($R^2$). 
We utilize NVIDIA Titan RTX GPU to train our models.
Tables~\ref{table:Pollution results}, ~\ref{table:Building results} and ~\ref{table:EHR results} presents the empirical results for the pollution, building and EHR datasets respectively.
The training time per epoch (the number of seconds to train a model once for the entire training set) is provided for each model, except for SVR-RBF, where the total training time is shown.
Unlike the building dataset, SVR-RBF requires many more iterations to reach convergence within the acceptable error threshold ($\epsilon = 0.1$) for both pollution and EHR datasets resulting in much higher training and testing times.
Tables~\ref{table:Pollution results}, ~\ref{table:Building results} and ~\ref{table:EHR results} also show that STAM is scalable and computationally tractable with comparable training time to DA-RNN. 
In terms of prediction performance, STAM maintains high accuracy and even slightly outperforms the baseline models in all three datasets. Attention-based time series models achieving comparable performance to that of the baseline models have also been reported previously \cite{choi2016retain, singh2017attend}. 
While maintaining high accuracy, STAM also provides accurate spatiotemporal interpretability.

\subsection{Discussion on Interpretability}
The proposed STAM model has been empirically shown to outperform the baseline models. 
As attention-based models typically experience a trade-off between prediction accuracy and interpretability, we next discuss how the proposed approach enables accurate spatiotemporal interpretability. 

{
\setlength{\tabcolsep}{0.5pt}
\begin{table*}[h]
\centering
\caption{\textit{Spatial attention weight distributions from STAM and DA-RNN.}}
\begin{tabular}{|c|c|c||c|c|c||c|c|c|}
\hline
\multicolumn{3}{|c||}{Pollution Dataset} & \multicolumn{3}{c||}{Building Dataset} & \multicolumn{3}{c|}{EHR Dataset} \\ 
\hline
\multirow{3}{*}{Variables} & \multicolumn{2}{|c||}{Attention} & \multirow{3}{*}{Variables} & \multicolumn{2}{|c||}{Attention} & \multirow{3}{*}{Variables} & \multicolumn{2}{|c|}{Attention} \\ 
& \multicolumn{2}{|c||}{Weight (\%)} & & \multicolumn{2}{|c||}{Weight (\%)} & & \multicolumn{2}{|c|}{Weight (\%)}\\
\cline{2-3}\cline{5-6}\cline{8-9}
 & STAM & DA-RNN & & STAM & DA-RNN & & STAM & DA-RNN\\ \hline\hline
\textbf{Pollution} & \textbf{13.43} & \textbf{13.26} & \textbf{AvgZoneTemp} & \textbf{11.76} & \textbf{11.22} & \textbf{Glucose} & \textbf{11.49} & \textbf{20.59}\\ \hline
\textbf{Dew Point} & \textbf{11.02} & 9.75 & OAT & 7.02 & 10.88 & Heart Rate & 5.74 & 9.93\\ 
\hline
Temperature & 10.64 & 9.93 & RAT & 7.48 & 10.99 & MBP & 8.64 & 12.64\\ \hline
Pressure & 10.81 & 10.30 &\textbf{OA DamperCMD} & \textbf{20.26} & \textbf{11.53} & $O_2$sat & 4.67 & 4.46\\ \hline
Wind Direction & 10.42 & 10.74 & CoolValveCMD & 9.13 & 11.00 & \textbf{RR} & \textbf{47.36} & \textbf{21.00}\\ \hline
\textbf{Wind Speed} & \textbf{14.28} & \textbf{14.91} & \textbf{DAT} & \textbf{14.73} & \textbf{11.32} & Temp. & 5.13 & 5.59\\ 
\hline
\textbf{Hours of Snow} & \textbf{14.68} & \textbf{15.55} & SuFanSpeedCMD & 8.99 & 10.90 & \textbf{pH} & \textbf{16.91} & \textbf{25.74}\\ 
\hline
\textbf{Hours of Rain} & \textbf{14.64} & \textbf{15.49} & DA StaticP & 9.28 & 10.96 & & &\\ \hline
 &  & & \textbf{ReFanSpeedCMD} & \textbf{11.26} & \textbf{11.12} & & &\\ \hline
\end{tabular}
\label{table:Weight Table}
\end{table*}
}

\begin{figure*}[ht]
  \centering
      \includegraphics[width=13.5cm]{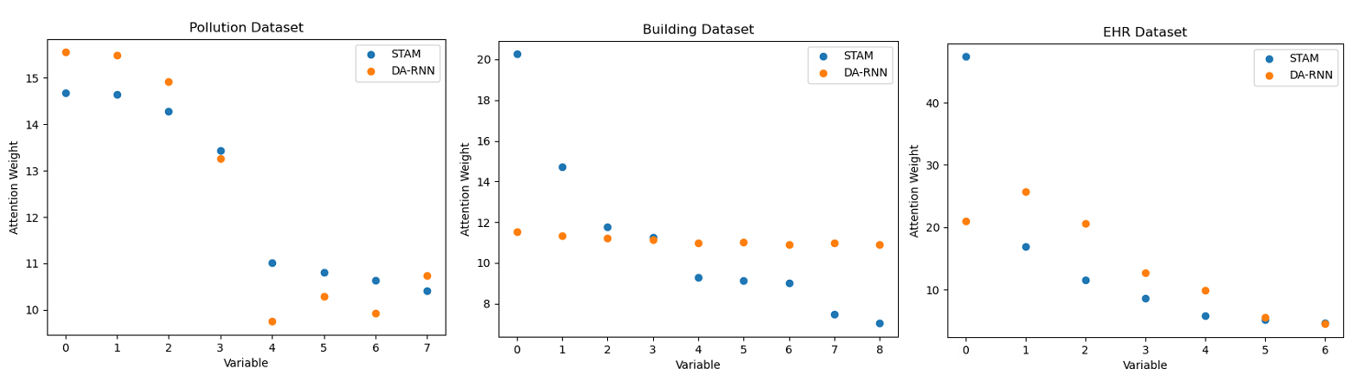} 
  \caption{\textit{Plots of spatial attention weight distributions from STAM and DA-RNN for pollution, building and EHR datasets. The variables are organized based on attention weights of STAM (high to low).}}
  \vspace{-7pt}
  \label{stam_darnn_spatial_weights}
\end{figure*}

\medskip
\noindent\textbf{Pollution Dataset:} 
When the output variable at a future time-step $T_y$ is \textit{pollution}, the \textit{temporal attention} weights are concentrated mostly in the last few time-steps (hours) of the input sequence decreasing from $T_x=5$ to $T_x=1$. Intuitively, the output in a time series is also affected mostly by recent time-steps, and the correlation usually decreases with increasing time lag.
From Table~\ref{table:Weight Table}, according to the spatial attention weights, the five most relevant variables from STAM include pollution, dew point, wind speed, hours of snow, and hours of rain. Temperature, pressure, and wind direction have comparatively smaller weights.
Such spatial correlations obtained by STAM are in accord with the specific underlying relationships found in~\cite{qi2018deep}. 
Intuitively, wind direction and pressure are not supposed to have strong correlations with pollution as PM2.5 refers to the atmospheric particulate matters which could be cleaned by weather conditions of snow or rain and blown away depending on the wind strength (speed) rather than direction. 
Dew point has a slightly higher attention weight than temperature(air) as dew point approximately shows how moist the air would be, which can affect the movement of the atmospheric particles.
From Table~\ref{table:Weight Table} and the plot in Fig.~\ref{stam_darnn_spatial_weights}, we observe that similar spatial interpretations are generated by both STAM and DA-RNN, while the slight differences are attributed to the different underlying model architectures.

\medskip
\noindent\textbf{Building Dataset:} 
With the \textit{average zone temperature} (AvgZoneTemp) as the output variable, the \textit{temporal attention} weights are almost equally distributed across all the input time-steps, suggesting that most likely, the correlation is weakly depending on the time in this case with a relatively high sampling frequency (one minute). This is attributed to the slow thermal dynamics in the zone and the impact of the building mass absorbing heat to resist the quick change of the zone temperature. 
Table~\ref{table:Weight Table} shows that the most relevant variables found by STAM are outside air damper command (OA DamperCMD), discharge air temperature (DAT), return fan speed command (ReFanSpeedCMD), and itself. Such correlations can be interpreted well by the domain knowledge (detailed physics provided in the supplementary section).
First, the AvgZoneTemp is affected by DAT since these two variables have a direct physics relationship based on the heat transfer equation, and in most building systems, the discharge air is pumped to the zone directly in summer without reheating. 
Return air temperature (RAT) indicates the temperature of zone air circulated back to the central system, and due to the summer time of data collection, it is similar to the outside air temperature (OAT).
Since the return air has a relatively higher level of $\textnormal{CO}_2$, only part of it is mixed with the fresh outside air to generate the mixed air, which is then cooled down by the cooling valve to become discharge air. Thus, how much fresh outside air and return air is required to maintain the indoor comfort are determined by the OA DamperCMD and ReFanSpeedCMD, respectively, which significantly affects the AvgZoneTemp. The cooling valve command (CoolValveCMD) controlling the cooled water flow rate directly affects the mixed air temperature instead of zone temperature, resulting in a smaller attention weight.
The discharge air static pressure (DA StaticP) has more impact on the airflow rate than OAT and RAT. 
The supply fan speed command (SuFanSpeedCMD), a key indicator for the airflow rate, has its attention weight closer to DA StaticP. 
From Table~\ref{table:Weight Table} and Fig.~\ref{stam_darnn_spatial_weights}, DA-RNN fails to distinguish the most informative variables clearly with the weights learned roughly the same for all. It implies that it may not efficiently capture the underlying complex system dynamics. In contrast, STAM is much better at learning the system dynamics through the data associated with different variables.

\medskip
\noindent\textbf{EHR Dataset:} 
Researchers have concluded that heart rate (HR) does not have much impact on the respiratory rate (RR) \cite{sciencedaily2020, druschky2020effects}. Also, RR measurements correlate poorly with oxygen saturation ($O_2$sat) measurements \cite{mower1996comparison}. 
These two relations are precisely reflected from the interpretations of STAM in Table~\ref{table:Weight Table}, with HR and $O_2$sat having lower attention weights to predict RR. 
The \textit{temporal attention} weights are found to be almost uniformly distributed across the time-steps.
The most relevant variables found by STAM are glucose and pH apart from RR itself, as shown in Table~\ref{table:Weight Table}, which can be interpreted well.
Glucose levels can affect RR, and the risk of having difficulty in breathing can be reduced by maintaining good diabetes control \cite{diabetesuk2019}. Rapid or labored breathing can be a symptom of diabetic ketoacidosis (DKA), which is caused by very high blood glucose levels accompanied by a high level of ketones in the blood. DKA can have serious consequences on the respiratory system \cite{de2019effects}. pH describes the acidity or basicity of a solution. For good health, the right pH levels are needed, and the human body constantly works to control the pH level of blood. Disturbance in the acid-base balance of the body changes the pH level. It provokes automatic compensatory mechanisms to push the blood pH back towards normal.
In general, the respiratory system is utilized to compensate for metabolic disturbances \cite{merckmanuals2020}. Therefore, the pH level can significantly affect RR.
From Table~\ref{table:Weight Table} and the plot in Fig.~\ref{stam_darnn_spatial_weights}, STAM demonstrates better distinguishing capability than DA-RNN with comparable interpretations from both methods for RR prediction.

\section{Conclusion}
Multivariate time series prediction has numerous applications in various domains of significant societal impact such as healthcare, financial markets, agriculture and climate science to name a few. Performance monitoring of human-engineered systems also utilizes multivariate sensor time series data to a great extent for enhanced efficiency, safety and security.
In this paper, we propose a deep learning architecture that can effectively capture the temporal correlations in multivariate time series data and simultaneously provide accurate interpretations of the prediction outcomes. 
The temporal and spatial attentions are directly aligned with the output variable.
Leveraging the spatial context vector facilitates better spatial correlation-based interpretations between the input and output variables.

Such spatial and temporal interpretations can help users to better understand the contributions of different features and time-steps for prediction. Therefore, we believe that the work presented here can be a great resource for the domain experts in different sectors who seek trustworthy (not just a black-box) machine learning tool for analyzing time series data. Scientists, engineers, medical professionals, farmers, policy makers and various other types of users can gain valuable insights by understanding the predictions from the perspective of 'what' and 'where' highlighted by our proposed model. Therefore, the proposed model, which is applicable to any type of multivariate time series data can help in data-driven decision-making in various parts of the society, well-beyond the machine learning community.

While the spatial and temporal correlations are captured based on the attention weights, one potential limitation is how much data is required to achieve it, which remains an open problem in this work. The required number of data points could depend on how prominent the underlying relationships are between the input and output variables in a dataset.  
In future, we plan to extend STAM to problems involving multiple output variables.


\bibliographystyle{unsrt}
\bibliography{neurips_2020.bib}

\begin{thebibliography}{10}

\bibitem{bahadori2019temporal}
Mohammad~Taha Bahadori and Zachary~Chase Lipton.
\newblock Temporal-clustering invariance in irregular healthcare time series.
\newblock {\em arXiv preprint arXiv:1904.12206}, 2019.

\bibitem{d2019trimmed}
Pierpaolo D’Urso, Livia De~Giovanni, and Riccardo Massari.
\newblock Trimmed fuzzy clustering of financial time series based on dynamic
  time warping.
\newblock {\em Annals of Operations Research}, pages 1--17, 2019.

\bibitem{mudelsee2019trend}
Manfred Mudelsee.
\newblock Trend analysis of climate time series: A review of methods.
\newblock {\em Earth-science reviews}, 190:310--322, 2019.

\bibitem{abatzoglou2020multivariate}
John~T Abatzoglou, Solomon~Z Dobrowski, and Sean~A Parks.
\newblock Multivariate climate departures have outpaced univariate changes
  across global lands.
\newblock {\em Scientific reports}, 10(1):1--9, 2020.

\bibitem{gonzalez2019methodology}
Aurora Gonzalez-Vidal, Fernando Jimenez, and Antonio~F Gomez-Skarmeta.
\newblock A methodology for energy multivariate time series forecasting in
  smart buildings based on feature selection.
\newblock {\em Energy and Buildings}, 196:71--82, 2019.

\bibitem{zhang2019deep}
Chuxu Zhang, Dongjin Song, Yuncong Chen, Xinyang Feng, Cristian Lumezanu, Wei
  Cheng, Jingchao Ni, Bo~Zong, Haifeng Chen, and Nitesh~V Chawla.
\newblock A deep neural network for unsupervised anomaly detection and
  diagnosis in multivariate time series data.
\newblock In {\em Proceedings of the AAAI Conference on Artificial
  Intelligence}, volume~33, pages 1409--1416, 2019.

\bibitem{malhotra2015long}
Pankaj Malhotra, Lovekesh Vig, Gautam Shroff, and Puneet Agarwal.
\newblock Long short term memory networks for anomaly detection in time series.
\newblock In {\em European Symposium on Artificial Neural Networks,
  Computational Intelligence and Machine Learning. Bruges (Belgium)}, page~89.
  Presses universitaires de Louvain, 2015 April 22-24.

\bibitem{shook2018integrating}
Johnathon Shook, Linjiang Wu, Tryambak Gangopadhyay, Baskar
  Ganapathysubramanian, Soumik Sarkar, and Asheesh~K Singh.
\newblock Integrating genotype and weather variables for soybean yield
  prediction using deep learning.
\newblock {\em bioRxiv}, 2018.

\bibitem{hua2019deep}
Yuxiu Hua, Zhifeng Zhao, Rongpeng Li, Xianfu Chen, Zhiming Liu, and Honggang
  Zhang.
\newblock Deep learning with long short-term memory for time series prediction.
\newblock {\em IEEE Communications Magazine}, 57(6):114--119, 2019.

\bibitem{gangopadhyay2020deep}
Tryambak Gangopadhyay, Anthony Locurto, James~B Michael, and Soumik Sarkar.
\newblock Deep learning algorithms for detecting combustion instabilities.
\newblock In {\em Dynamics and Control of Energy Systems}, pages 283--300.
  Springer, 2020.

\bibitem{cho2014learning}
Kyunghyun Cho, Bart Van~Merri{\"e}nboer, Caglar Gulcehre, Dzmitry Bahdanau,
  Fethi Bougares, Holger Schwenk, and Yoshua Bengio.
\newblock Learning phrase representations using rnn encoder-decoder for
  statistical machine translation.
\newblock {\em arXiv preprint arXiv:1406.1078}, 2014.

\bibitem{sutskever2014sequence}
Ilya Sutskever, Oriol Vinyals, and Quoc~V Le.
\newblock Sequence to sequence learning with neural networks.
\newblock In {\em Advances in neural information processing systems}, pages
  3104--3112, 2014.

\bibitem{sagheer2019unsupervised}
Alaa Sagheer and Mostafa Kotb.
\newblock Unsupervised pre-training of a deep lstm-based stacked autoencoder
  for multivariate time series forecasting problems.
\newblock {\em Scientific Reports}, 9(1):1--16, 2019.

\bibitem{bahdanau2014neural}
Dzmitry Bahdanau, Kyunghyun Cho, and Yoshua Bengio.
\newblock Neural machine translation by jointly learning to align and
  translate.
\newblock {\em arXiv preprint arXiv:1409.0473}, 2014.

\bibitem{choi2016retain}
Edward Choi, Mohammad~Taha Bahadori, Jimeng Sun, Joshua Kulas, Andy Schuetz,
  and Walter Stewart.
\newblock Retain: An interpretable predictive model for healthcare using
  reverse time attention mechanism.
\newblock In {\em Advances in Neural Information Processing Systems}, pages
  3504--3512, 2016.

\bibitem{qin2017dual}
Yao Qin, Dongjin Song, Haifeng Chen, Wei Cheng, Guofei Jiang, and Garrison
  Cottrell.
\newblock A dual-stage attention-based recurrent neural network for time series
  prediction.
\newblock {\em arXiv preprint arXiv:1704.02971}, 2017.

\bibitem{singh2017attend}
Ritambhara Singh, Jack Lanchantin, Arshdeep Sekhon, and Yanjun Qi.
\newblock Attend and predict: Understanding gene regulation by selective
  attention on chromatin.
\newblock In {\em Advances in neural information processing systems}, pages
  6785--6795, 2017.

\bibitem{song2018attend}
Huan Song, Deepta Rajan, Jayaraman~J Thiagarajan, and Andreas Spanias.
\newblock Attend and diagnose: Clinical time series analysis using attention
  models.
\newblock In {\em Thirty-second AAAI conference on artificial intelligence},
  2018.

\bibitem{gangopadhyay2018temporal}
Tryambak Gangopadhyay, Sin~Yong Tan, Genyi Huang, and Soumik Sarkar.
\newblock Temporal attention and stacked lstms for multivariate time series
  prediction.
\newblock In {\em NeurIPS Workshop on Modeling and Decision-Making in the
  Spatiotemporal Domain}. NeurIPS, 2018.

\bibitem{zhang2019lstm}
Xuan Zhang, Xun Liang, Aakas Zhiyuli, Shusen Zhang, Rui Xu, and Bo~Wu.
\newblock At-lstm: An attention-based lstm model for financial time series
  prediction.
\newblock In {\em IOP Conference Series: Materials Science and Engineering},
  volume 569, page 052037. IOP Publishing, 2019.

\bibitem{li2019ea}
Youru Li, Zhenfeng Zhu, Deqiang Kong, Hua Han, and Yao Zhao.
\newblock Ea-lstm: Evolutionary attention-based lstm for time series
  prediction.
\newblock {\em Knowledge-Based Systems}, 181:104785, 2019.

\bibitem{liu2020dstp}
Yeqi Liu, Chuanyang Gong, Ling Yang, and Yingyi Chen.
\newblock Dstp-rnn: A dual-stage two-phase attention-based recurrent neural
  network for long-term and multivariate time series prediction.
\newblock {\em Expert Systems with Applications}, 143:113082, 2020.

\bibitem{shook2020crop}
Johnathon Shook, Tryambak Gangopadhyay, Linjiang Wu, Baskar
  Ganapathysubramanian, Soumik Sarkar, and Asheesh~K Singh.
\newblock Crop yield prediction integrating genotype and weather variables
  using deep learning.
\newblock {\em arXiv preprint arXiv:2006.13847}, 2020.

\bibitem{ismail2019input}
Aya~Abdelsalam Ismail, Mohamed Gunady, Luiz Pessoa, Hector~Corrada Bravo, and
  Soheil Feizi.
\newblock Input-cell attention reduces vanishing saliency of recurrent neural
  networks.
\newblock In {\em Advances in Neural Information Processing Systems}, pages
  10813--10823, 2019.

\bibitem{vaswani2017attention}
Ashish Vaswani, Noam Shazeer, Niki Parmar, Jakob Uszkoreit, Llion Jones,
  Aidan~N Gomez, {\L}ukasz Kaiser, and Illia Polosukhin.
\newblock Attention is all you need.
\newblock In {\em Advances in neural information processing systems}, pages
  5998--6008, 2017.

\bibitem{hu2020multistage}
Jun Hu and Wendong Zheng.
\newblock Multistage attention network for multivariate time series prediction.
\newblock {\em Neurocomputing}, 383:122--137, 2020.

\bibitem{du2020multivariate}
Shengdong Du, Tianrui Li, Yan Yang, and Shi-Jinn Horng.
\newblock Multivariate time series forecasting via attention-based
  encoder-decoder framework.
\newblock {\em Neurocomputing}, 2020.

\bibitem{zhang2018recurrent}
Mingxing Zhang, Yang Yang, Yanli Ji, Ning Xie, and Fumin Shen.
\newblock Recurrent attention network using spatial-temporal relations for
  action recognition.
\newblock {\em Signal Processing}, 145:137--145, 2018.

\bibitem{wang2016attention}
Yequan Wang, Minlie Huang, Xiaoyan Zhu, and Li~Zhao.
\newblock Attention-based lstm for aspect-level sentiment classification.
\newblock In {\em Proceedings of the 2016 conference on empirical methods in
  natural language processing}, pages 606--615, 2016.

\bibitem{hochreiter1997long}
Sepp Hochreiter and J{\"u}rgen Schmidhuber.
\newblock Long short-term memory.
\newblock {\em Neural computation}, 9(8):1735--1780, 1997.

\bibitem{yoshimiforecasting}
Shuhei Yoshimi and Koji Eguchi.
\newblock Forecasting corporate financial time series using multi-phase
  attention recurrent neural networks.
\newblock 2020.

\bibitem{tran2018temporal}
Dat~Thanh Tran, Alexandros Iosifidis, Juho Kanniainen, and Moncef Gabbouj.
\newblock Temporal attention-augmented bilinear network for financial
  time-series data analysis.
\newblock {\em IEEE transactions on neural networks and learning systems},
  30(5):1407--1418, 2018.

\bibitem{liang2015assessing}
Xuan Liang, Tao Zou, Bin Guo, Shuo Li, Haozhe Zhang, Shuyi Zhang, Hui Huang,
  and Song~Xi Chen.
\newblock Assessing beijing's pm2. 5 pollution: severity, weather impact, apec
  and winter heating.
\newblock {\em Proc. R. Soc. A}, 471(2182):20150257, 2015.

\bibitem{OpenEI}
U.S.~Department of~Energy~(DOE).
\newblock Long-term data on 3 office air handling units.
\newblock
  \url{http://aiweb.techfak.uni-bielefeld.de/content/bworld-robot-control-software/},
  July 21, 2015.

\bibitem{johnson2016mimic}
Alistair~EW Johnson, Tom~J Pollard, Lu~Shen, H~Lehman Li-Wei, Mengling Feng,
  Mohammad Ghassemi, Benjamin Moody, Peter Szolovits, Leo~Anthony Celi, and
  Roger~G Mark.
\newblock Mimic-iii, a freely accessible critical care database.
\newblock {\em Scientific data}, 3(1):1--9, 2016.

\bibitem{harutyunyan2019multitask}
Hrayr Harutyunyan, Hrant Khachatrian, David~C Kale, Greg Ver~Steeg, and Aram
  Galstyan.
\newblock Multitask learning and benchmarking with clinical time series data.
\newblock {\em Scientific data}, 6(1):1--18, 2019.

\bibitem{qi2018deep}
Zhongang Qi, Tianchun Wang, Guojie Song, Weisong Hu, Xi~Li, and Zhongfei Zhang.
\newblock Deep air learning: Interpolation, prediction, and feature analysis of
  fine-grained air quality.
\newblock {\em IEEE Transactions on Knowledge and Data Engineering},
  30(12):2285--2297, 2018.

\bibitem{sciencedaily2020}
ScienceDaily.
\newblock {\em American Institute Of Physics. Heartbeat And Breathing Cycles.},
  2007.

\bibitem{druschky2020effects}
Katrin Druschky, J{\"u}rgen Lorenz, and Achim Druschky.
\newblock Effects of respiratory rate on heart rate variability in neurologic
  outpatients with epilepsies or migraine: A preliminary study.
\newblock {\em Medical Principles and Practice}, 29(4):318--325, 2020.

\bibitem{mower1996comparison}
WR~Mower, C~Sachs, EL~Nicklin, P~Safa, and LJ~Baraff.
\newblock A comparison of pulse oximetry and respiratory rate in patient
  screening.
\newblock {\em Respiratory medicine}, 90(10):593--599, 1996.

\bibitem{diabetesuk2019}
Diabetes.co.uk.
\newblock {\em Respiratory System and Diabetes.}, 2019.

\bibitem{de2019effects}
Alice~Gallo de~Moraes and Salim Surani.
\newblock Effects of diabetic ketoacidosis in the respiratory system.
\newblock {\em World Journal of Diabetes}, 10(1):16, 2019.

\bibitem{merckmanuals2020}
Merck Manuals.
\newblock {\em Overview of Acid-Base Balance.}, 2020.

\bibitem{charles2013adoption}
Dustin Charles, Meghan Gabriel, and Michael~F Furukawa.
\newblock Adoption of electronic health record systems among us non-federal
  acute care hospitals: 2008-2012.
\newblock {\em ONC data brief}, 9:1--9, 2013.

\end{thebibliography}

\newpage

\section*{Supplementary Materials}

\section*{S.1 Baseline Models}
We use the following baseline models to compare the results with our proposed models STAM and STAM-Lite.
We try to optimize the hyper-parameters of the baseline models.

\begin{enumerate}

    \item \textbf{SVR-RBF:}
    Epsilon-Support Vector Regression with Radial Basis Function kernel. We optimize the hyper-parameters of SVR-RBF, especially epsilon. 
    After choosing the optimal set of hyperparameters, we observe that SVR-RBF shows improved performance than in \cite{du2020multivariate} for the pollution dataset.
    
    \item \textbf{Enc-Dec:}
    Encoder-Decoder model \cite{cho2014learning, sutskever2014sequence}. Originally developed for neural machine translation, a deep LSTM based encoder-decoder model has also been applied for multivariate time series prediction \cite{sagheer2019unsupervised}.
    The Enc-Dec model comprises an encoder for the input sequence and decoder for the output sequence. The encoder encodes the input sequence into a fixed-length vector used by the decoder to predict the output sequence.
    The whole encoder-decoder setup is trained jointly.

    \item \textbf{LSTM-Att:} 
    LSTM with temporal attention model.
    Proposed as an extension to the Encoder-Decoder model \cite{cho2014learning, sutskever2014sequence}, the attention-based model \cite{bahdanau2014neural} can automatically soft search for important parts of the input sequence. Instead of encoding the whole input sequence into a single fixed-length vector, the model encodes the input sequence into a sequence of vectors. During decoding, it adaptively chooses a subset of these vectors where the most relevant information is concentrated. Intuitively, a similar approach like this has been adopted for temporal attention in time series prediction \cite{singh2017attend, du2020multivariate}. 

    Both the \textit{Enc-Dec} and \textit{LSTM-Att} models were originally developed for neural machine translation. For time series prediction, the output at each decoder time-step is a scalar instead of a vector. We modify both of these models accordingly using mean squared error as the loss. Similar to STAM-Lite and STAM, the decoder receives the information of the previously predicted output instead of the real measurement through the non-teacher forcing training approach. 

    \item \textbf{DA-RNN:} 
    Dual-stage Attention-based Recurrent Neural Networks \cite{qin2017dual}. It has spatial attention in the encoder layer, which computes a sequence of hidden states. As described in our main paper content, one of the limitations of DA-RNN is that it is not causal, depending on the future inputs during the encoding phase. The temporal attention weights are computed in the decoding phase.   
    
\end{enumerate}

\begin{table*}[ht]
\caption{\textit{Approximate number of trainable parameters with input sequence length, $T_x=5$ and output sequence length, $T_y=3$ for each baseline model (except SVR-RBF) and our proposed STAM-Lite and STAM models.}}
\begin{center}
\begin{threeparttable}
\begin{tabular}{|l||c|c|c|c|c|}
    \toprule
Model & Enc-Dec & LSTM-Att & DA-RNN & STAM-Lite & STAM\\ \midrule\midrule
Number of Parameters & $\sim$18,831 & $\sim$19,120 & $\sim$57,764 & $\sim$19,761 & $\sim$24,733 \\
    \bottomrule
\end{tabular}
\end{threeparttable}
\end{center}
\label{table:Number of Parameters}
\end{table*}

\section*{S.2 STAM-Lite}
\begin{figure*}
  \centering
  \includegraphics[width=13.5cm,height=7cm]{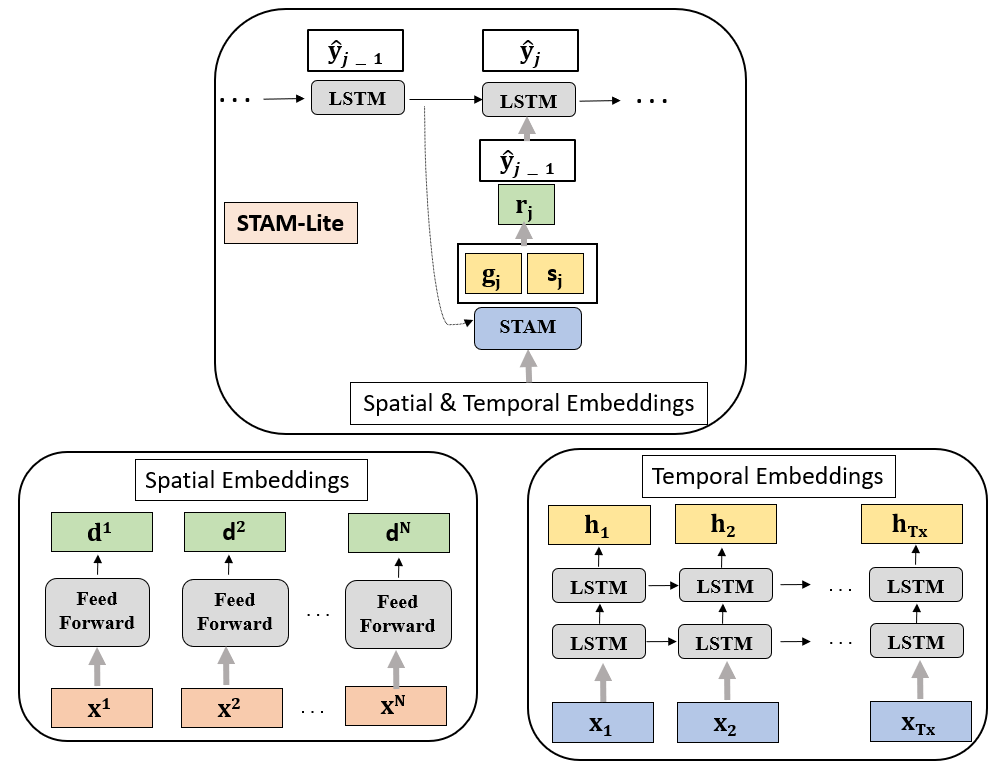} 
  \caption{\textit{The illustration of the proposed model STAM-Lite attempting to compute the output $\hat{\mathbf{y}}_j$ at time step $j$.}}
  \vspace{-7pt}
  \label{stam_1}
\end{figure*}

In STAM-Lite, there is a single LSTM layer in the decoder.  At output time-step $j$, the context vectors $\mathbf{g}_j$ and $\mathbf{s}_j$ are computed using the previous decoder hidden state $\mathbf{h'}_{j-1}$. To align both the contexts with the output time series, we first concatenate these two vectors into $[\mathbf{g}_j;\mathbf{s}_j]\in\mathbb{R}^{2m}$.
The concatenated dimension is reduced to $\mathbf{r}_j\in\mathbb{R}^q$ using a feed-forward neural network, as shown in Eq.~\ref{stam1_concat}. We optimize the extent of this reduction through our experiments. Next we update $\mathbf{r}_j$ by concatenating with output of the previous time-step $\hat{\mathbf{y}}_{j-1}$. The concatenation is denoted by $\hat{\mathbf{r}}_j$. It should be noted that the decoder output $\hat{\mathbf{y}}_{j-1}$ is a scalar in time series prediction instead of a vector.
\begin{equation}\label{stam1_concat}
    \begin{split}
    &\mathbf{r}_j = \textnormal{ReLU}(W_{GS}[\mathbf{g}_j;\mathbf{s}_j]+b_{GS}),\;
    \hat{\mathbf{r}}_j = [\mathbf{r}_j; \hat{\mathbf{y}}_{j-1}]
    \end{split}
\end{equation}
where $W_{GS}\in\mathbb{R}^{q\times 2m}$ and $b_{GS}\in\mathbb{R}^q$ are parameters to learn. Instead of the real measurement, the prediction of the output time series is utilized through \textit{non-teacher forcing} training approach, which enables the learning to be more robust. The decoder hidden state $\mathbf{h'}_j$ and cell state $\mathbf{h'}_j$ are updated by using $\hat{\mathbf{r}}_j$ as input to the decoder LSTM (function $f_2$):
\begin{equation}\label{stam1_dec_hidden}
    (\mathbf{h'}_{j}, \mathbf{c'}_{j}) = f_2(\mathbf{h'}_{j-1}, \mathbf{c'}_{j-1}, \hat{\mathbf{r}}_j).    
\end{equation}
From Fig.~\ref{stam_1}, we can observe that the spatiotemporal context vector is aligned directly with the output variable enabling accurate interpretability. Eq.~\ref{stam1_dec_hidden} suggests that in STAM-Lite, both the spatial and temporal attentions share the same decoder parameterized by an LSTM layer, which is due to the concatenation of the two context vectors.

\subsection*{S.2.1 Complexity Analysis}
STAM-Lite involves four modules: the encoder, the spatial attention, the temporal attention, and the decoder. Similarly, we still omit some operations such as spatial embeddings and concatenation reduction for the context vectors. 
The encoders and decoders in both STAM-Lite are spanned by LSTM layers, and it has two LSTM layers in the encoder.
We analyze the inference time complexity of the STAM-Lite model next. 
We still follow a similar way of the calculation as in \cite{tran2018temporal}. 
Since the state size of the encoder is $m$ and that of the decoder is $p$, such a mechanism leads to the inference time complexity of $\mathcal{O}(8(Nm+m^2+2m)T_x+(p+2+2m)(N+T_x)T_y+4T_y(p^2+pq+3p))$, where $q$ is the dimension of the context vector after dimension reduction. The first and third terms inside signify the computational complexity for the encoder and decoder, respectively. The second term is for both the spatial and temporal attention.
Thus, compared to STAM, STAM-Lite has a smaller complexity due to only one LSTM layer in the decoder module. 

\section*{S.3 Results}
We present here some additional results for pollution, building, and EHR datasets.

\begin{table*}[ht]
\caption{\textit{Empirical results for \textbf{pollution} dataset (with $T_x=5$, $T_y=3$). Each model was trained three times, to obtain the average and standard deviation of each evaluation metric.}}
\begin{center}
\begin{threeparttable}
\begin{tabular}{|l|c|c|c|}
    \toprule
Model     & RMSE                      & MAE                       & $R^2$ Score\\\midrule\midrule
SVR-RBF & 42.399 $\pm$ 0.000 & 27.917 $\pm$ 0.000 & 0.795 $\pm$ 0.000\\ \midrule
Enc-Dec & 42.105 $\pm$ 0.155        & 25.432 $\pm$  0.477        & 0.798 $\pm$  0.002\\ \midrule
LSTM-Att  & 41.739 $\pm$ 0.564          & \textbf{25.683 $\pm$ 0.969} & 0.801 $\pm$ 0.005\\ \midrule
DA-RNN  & 43.186 $\pm$ 0.051 & 26.455 $\pm$ 0.224 & 0.787 $\pm$ 0.001\\ \midrule
STAM-Lite    & 41.873 $\pm$ 0.200          & 25.851 $\pm$ 0.785          & 0.800 $\pm$ 0.002\\ \midrule
STAM    & \textbf{41.535 $\pm$ 0.216} & 25.691 $\pm$ 0.738          & \textbf{0.803 $\pm$ 0.002}\\
      \bottomrule
\end{tabular}
\end{threeparttable}
\end{center}
\label{table:Pollution results supp}
\end{table*}

\begin{table*}[ht]
\caption{\textit{Empirical results for \textbf{building} dataset (with $T_x=5$, $T_y=2$). Each model was trained three times, to obtain the average and standard deviation of each evaluation metric.}}
\begin{center}
\begin{threeparttable}
\begin{tabular}{|l|c|c|c|}
    \toprule
 Model     & RMSE                      & MAE                       & $R^2$ Score\\\midrule\midrule
SVR-RBF & 0.1354 $\pm$ 0.0000 & 0.1127 $\pm$ 0.0000 & 0.9872 $\pm$ 0.0000\\ \midrule
Enc-Dec & 0.0616 $\pm$ 0.0061          & 0.0464 $\pm$ 0.0057          & 0.9973 $\pm$ 0.0005\\ \midrule
LSTM-Att  & 0.0554 $\pm$ 0.0024          & 0.0392 $\pm$ 0.0025          & 0.9979 $\pm$ 0.0002\\ \midrule
DA-RNN  & 0.0548 $\pm$ 0.0018	& 0.0400 $\pm$ 0.0023	& 0.9979 $\pm$ 0.0001\\ \midrule
STAM-Lite    & 0.0634 $\pm$ 0.0018          & 0.0491 $\pm$ 0.0026          & 0.9972 $\pm$ 0.0002\\ \midrule
STAM    & \textbf{0.0528 $\pm$ 0.0027} & \textbf{0.0367 $\pm$ 0.0024} & \textbf{0.9980 $\pm$ 0.0002}\\
      \bottomrule
\end{tabular}
\end{threeparttable}
\end{center}
\label{table:Building results supp}
\end{table*}

\begin{table*}[ht]
\caption{\textit{Empirical results for \textbf{EHR} dataset (with $T_x=24$, $T_y=2$). Each model was trained three times, to obtain the average and standard deviation of each evaluation metric.}}
\begin{center}
\begin{threeparttable}
\begin{tabular}{|l|c|c|c|c|c|}
    \toprule
Model     & RMSE                      & MAE                       & $R^2$ Score\\
 & & & \\\midrule\midrule
SVR-RBF & 4.3962 $\pm$ 0.0000 & 3.1510 $\pm$ 0.0000 & 0.4538 $\pm$ 0.0000\\\midrule
Enc-Dec & 4.3983 $\pm$ 0.0206 & 3.1670 $\pm$ 0.0290 & 0.4533 $\pm$ 0.0051\\\midrule
LSTM-Att  & 4.4083 $\pm$ 0.0208 & 3.1667 $\pm$ 0.0176 & 0.4508 $\pm$ 0.0052\\\midrule
DA-RNN  & 4.4628 $\pm$ 0.0396 & 3.1665 $\pm$ 0.0201 & 0.4372 $\pm$ 0.0100\\\midrule
STAM-Lite    & 4.3993 $\pm$ 0.0288 & 3.1637 $\pm$ 0.0283 & 0.4531 $\pm$ 0.0072\\\midrule
STAM & \textbf{4.3860 $\pm$ 0.0146} & \textbf{3.1277 $\pm$ 0.0096} & \textbf{0.4564 $\pm$ 0.0036}\\
      \bottomrule
\end{tabular}
\end{threeparttable}
\end{center}
\label{table:EHR results 1}
\end{table*}

\section*{S.4 Overview of the Building Dataset}
In this section, we provide an overview of the working mechanism of airside heating, ventilation, and air-conditioning (HVAC) in the building.
Every building HVAC system can be roughly divided into two sub-systems: airside HVAC system and waterside HVAC system. As the data used in our experiments is more related to the airside HVAC system, we omit the working mechanism of the waterside HVAC system in this context.

Regarding the airside HVAC system, although it is essentially interacted with the waterside HVAC system, in terms of energy, it is on the demand side, which means that according to each zone’s comfort requirement, the whole system generates the conditioned air to be supplied to the zones for satisfying the specific needs. We focus on the physical relationships qualitatively among the variables of the building dataset used in our experiments. We start with the definition of each variable.

\subsection*{S.4.1 Variables}

\begin{enumerate}

\item \textbf{Average Zone Temperature (Avg Zone Temp, $\degree F$):}
The temperature measured in an averaged way for a zone. This is because the local variable air volume supplying air to a zone can be located in different places inside the zone, depending on a specific building configuration. Thus, temperatures measured for different places inside the zone can be diverse. However, when collecting data, only one temperature is reported by averaging several slightly different zone temperatures.

\item \textbf{Outside Air Temperature (OAT, $\degree F$):}
The temperature of the outdoor environment.

\item \textbf{Return Air Temperature (RAT, $\degree F$):}
The temperature of the air returning from the zone. 

\item \textbf{Outside Air Damper Command (OA Damper CMD):}
The command signal of the damper position controlling the outside air. It ranges from 0 – 100\%, where 0\% means that the damper is closed while 100\% means that the damper is fully open.

\item \textbf{Cooling Valve Command (Cool Valve CMD):}
The command signal of the valve position controlling the cooling water. It ranges from 0 – 100\%, where 0\% means that the valve is closed while 100\% means that the valve is fully open.

\item \textbf{Discharge Air Temperature (DAT, $\degree F$):}
The temperature of the discharge air, which is supplied to the zone.

\item \textbf{Supply Fan Speed Command (Su Fan Speed CMD):}
The command signal of the supply fan to control how fast it runs. The supply fan is for pumping the discharge air.

\item \textbf{Discharge Air Static Pressure (DA StaticP):}
The static pressure of the discharge air.

\item \textbf{Return Fan Speed Command (Re Fan Speed CMD):}
The command signal of the return fan to control how fast it runs. The return fan facilitates to circulate back the air to the central airside system.

\end{enumerate}

\subsection*{S.4.2 Working Mechanism of the Airside HVAC System}
Before introducing how the airside HVAC system works, we first mention two notations that will be used next. The first one is called the air-handling unit (AHU), and the second one is the aforementioned variable air volume (VAV). Typically, AHU generates the conditioned air for each zone. For each zone, its local VAV will reheat the conditioned air based on different comfort requirements and seasons (e.g., cooling season and heating season). As the data used in the experiments are collected from the cooling season, we focus only on cooling down the zone temperature. We ignore any heating for the zone here.

When the fresh outside air is taken into the system by the AHU, at the same time, part of the return air from the thermal zone(s) is circulated back to the AHU.
Part of the return air is pumped out of the system as the exhaust air. One could reuse the total return air, but it would cause the ventilation issue as the CO2 level keeps increasing. The reason for not using the refreshed outside air entirely is due to the energy-saving, particularly for the winter (heating season). The outside air and return air is mixed in the AHU to become the mixed air. 

Due to the cooling season, the mixed air temperature is relatively high such that it cannot be provided directly to the thermal zone(s). Then the cooling water comes to play. Before reaching the supply fan, the mixed air passes a coil system in the AHU. Since it is the cooling season, the cooling coil is activated to cool down the mixed air to become the discharge air. Hence, the discharge air is pumped by the supply fan to the thermal zone(s). Before entering each thermal zone, depending on the record of the thermostat, the reheat coil on the VAV is activated or not. In the cooling season, this is not usual but also relies on different climates. After the discharge air enters into the thermal zone, it helps regulate the zone temperature to the reference temperature set by the thermostat.  The air from the thermal zone(s) is pumped by the return air fan back to the central AHU system. This process is repeated in the airside HVAC system.

\section*{S.5 EHR Dataset}

In the United States, most hospitals use an electronic health record (EHR) system across different states \cite{charles2013adoption}.
With time, more hospitals are adopting digital health record systems. 
To make hospital data widely accessible to researchers, Medical Information Mart for Intensive Care (MIMIC-III) database was released \cite{johnson2016mimic}.
MIMIC-III is a publicly available EHR database that comprises information relating to patients admitted to critical care units at a large tertiary care hospital.
Clinical prediction benchmarks have been proposed using MIMIC-III data \cite{harutyunyan2019multitask}. 
The proposed benchmark tasks contain multivariate time series data for intensive care unit (ICU) stays covering clinical problems like in-hospital mortality prediction, forecasting length of stay (LOS), detecting physiologic decline, and phenotype classification.
For example, task LOS involves a regression at each time step, while in-hospital mortality risk is predicted once early in admission.
In this paper, instead of using the benchmark tasks, we formulate a new task that can be most suitable to test our proposed models.
After getting access to MIMIC-III, we only followed \cite{harutyunyan2019multitask} to generate the datasets for training, validation, and test sets. 

In the benchmark \cite{harutyunyan2019multitask}, the time series are re-sampled into regularly spaced intervals. They performed imputation of the missing values using the most recent measurement value if it exists or a pre-specified ``normal" value. After following the benchmark processing, the number of ICU stays in training, validation, and test sets are 14681, 3222, and 3236, respectively. 

We choose seven clinical variables - Glucose, Heart Rate, Mean Blood Pressure (MBP), Oxygen Saturation ($O_2$sat), Respiratory Rate (RR), Temperature, and pH level. Keeping these variables as input for $T_x = 24$ time-steps, we predict respiratory rate for $T_y$ future time-steps. In other words, based on the clinical variables on the first day (24 hours) of an ICU stay, we predict the respiratory rate for the upcoming $T_y$ hours.

\end{document}